\documentclass[10pt,twocolumn,letterpaper]{article}

\usepackage{iccv}
\usepackage{times}
\usepackage{epsfig}
\usepackage{graphicx}
\usepackage{amsmath}
\usepackage{amssymb}
\usepackage{amsthm}

\usepackage{enumitem}
\usepackage{tikz}
\usepackage{comment}

\usepackage{color}
\usepackage{colortbl}
\usepackage{multirow}
\usepackage{booktabs}
\usepackage{subfig}
\usepackage{subfloat}
\usepackage{mathtools}
\usepackage{dsfont}
\usepackage{bm}
\usepackage[T1]{fontenc}
\usepackage{authblk}
\usepackage{fancyhdr}
\usepackage[accsupp]{axessibility}  

\definecolor{mygray}{gray}{.9}

\usepackage[breaklinks=true,bookmarks=false]{hyperref}

\newtheorem{definition}{Definition}
\newtheorem{remark}{Remark}

\iccvfinalcopy 

\def\iccvPaperID{2931} 

\ificcvfinal\fi

\begin{document}

\title{Universal Domain Adaptation via Compressive Attention Matching}

\author[1]{Didi Zhu \textsuperscript{*}}
\author[2]{Yinchuan Li \textsuperscript{*}}
\author[1]{Junkun Yuan}
\author[1]{Zexi Li}
\author[1]{Kun Kuang \textsuperscript{$\dagger$}}
\author[1]{Chao Wu \textsuperscript{$\dagger$}}
\affil[1]{Zhejiang University}
\affil[2]{Huawei Noah’s Ark Lab}
\affil[1]{\tt\small \{didi\_zhu, yuanjk, zexi.li, kunkuang, chao.wu\}@zju.edu.cn}
\affil[2]{\tt\small \{liyinchuan\}@huawei.com}

\maketitle

\renewcommand*{\thefootnote}{}
\footnotetext{
* Equal contributions.}
\footnotetext{
$\dagger$ Corresponding authors.}
\renewcommand*{\thefootnote}{\arabic{footnote}}
\setcounter{footnote}{0}



\ificcvfinal\thispagestyle{empty}\fi

\begin{abstract}
  Universal domain adaptation (UniDA) aims to transfer knowledge from the source domain to the target domain without any prior knowledge about the label set. The challenge lies in how to determine whether the target samples belong to common categories. The mainstream methods make judgments based on the sample features, which overemphasizes global information while ignoring the most crucial local objects in the image, resulting in limited accuracy. To address this issue, we propose a Universal Attention Matching (UniAM) framework by exploiting the self-attention mechanism in vision transformer to capture the crucial object information. The proposed framework introduces a novel Compressive Attention Matching (CAM) approach to explore the core information by compressively representing attentions.  
  Furthermore, CAM incorporates a residual-based measurement to determine the sample commonness. By utilizing the measurement, UniAM achieves domain-wise and category-wise Common Feature Alignment (CFA) and Target Class Separation (TCS). Notably, UniAM is the first method utilizing the attention in vision transformer directly to perform classification tasks. Extensive experiments show that UniAM outperforms the current state-of-the-art methods on various benchmark datasets.

   
\end{abstract}

\section{Introduction\label{sec:intro}}
While deep neural networks have achieved remarkable success on visual tasks~\cite{deng2009large,krizhevsky2012imagenet,he2016deep,zhang2020knowledge,shen2020federated,zhang2022fairness,zhang2020federated}, their performance heavily relies on the assumption of independently and identically distributed (i.i.d.) training and test data~\cite{vapnik1991principles}. However, this assumption is frequently violated due to the presence of domain shift in real-world scenarios~\cite{shen2021towards,liu2021heterogeneous,liu2021kernelized, yuan2023instrumental,yuan2023domain,zhang2023rotogbml,zhang2022domain}.
Unsupervised Domain Adaptation (DA)~\cite{ben2010theory} has emerged as a promising solution to address this limitation by adapting models trained on a source domain to perform well on an unlabeled target domain. 
Nevertheless, most existing DA approaches~\cite{ganin2015unsupervised,tzeng2017adversarial,saito2018maximum,ma2022attention,lv2023duet,lv2023ideal} assume that the label spaces in the source and target domains are identical, which may not always hold in practical scenarios. 
Partial Domain Adaptation (PDA)~\cite{cao2018partial} and Open Set Domain Adaptation (OSDA)~\cite{panareda2017open} have been proposed to handle cases where the label spaces in one domain include those in the other, but these still rely on prior knowledge on label set, limiting knowledge generalizing from one scenario to others.
Universal domain adaptation (UniDA)~\cite{you2019universal} considers a more practical and challenging scenario where the relationship of label space between source and target domains is completely unknown
 i.e. with any number of common, source-private and target-private classes.

\begin{figure}[tbp]
	\centering
	\subfloat{
	\includegraphics[width=3.3in]{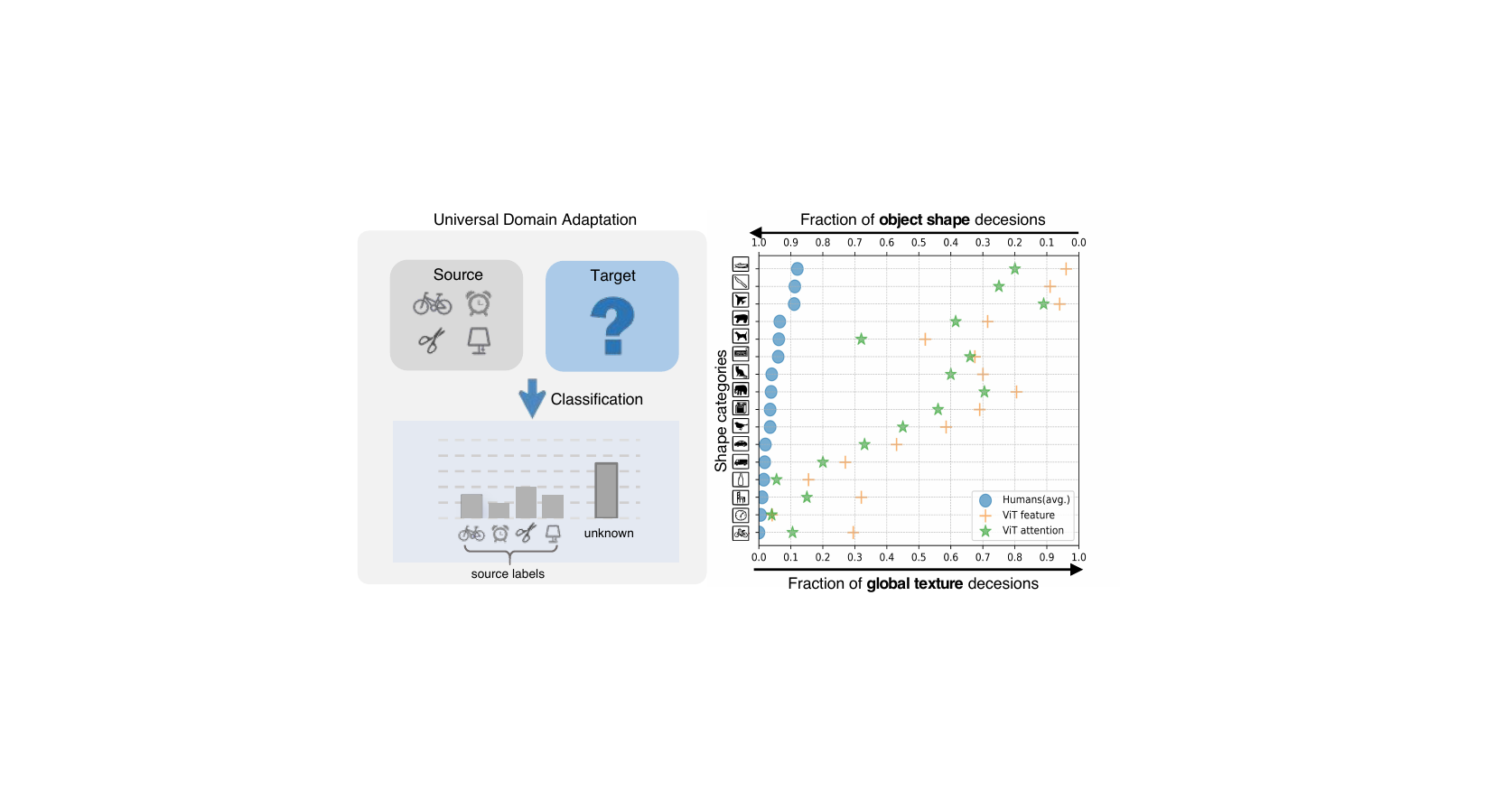}
	}
	\caption{\textbf{Left:} Illustration of Universal Domain Adaptation. \textbf{Right:} Shape-bias Analysis. Plot shows shape-texture trade off for attention and feature in ViT and humans.}
	\label{shape_texture}
\end{figure}

\begin{figure*}[htbp]
	\centering
	\subfloat{
	\includegraphics[width=6.8in]{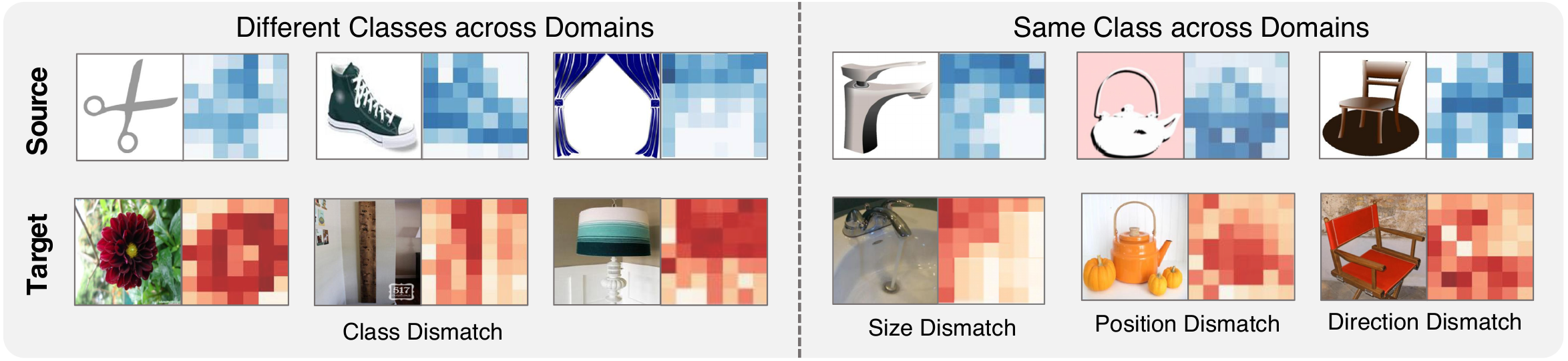}
	}
	\caption{\textbf{Attention Visualization accross domains.} 
    Attention patterns vary significantly between different classes of images. However, within the same class, attention can also exhibit variations due to differences in object size, position, and orientation. These variations are collectively referred to as \textit{attention mismatch}.}
	\label{attn_vis1}
\end{figure*}


In UniDA, the primary objective is to develop a model capable of precisely categorizing target samples as one of the common classes or an "unknown" class as shown in Fig.~\ref{shape_texture} left.
Existing UniDA methods aim to design a transferability criteria to detect common and private classes solely based on the discriminability of deep features~\cite{chang2022unified,chen2022mutual, chen2022evidential, chen2022geometric,fu2020learning,kundu2020universal,li2021domain,saito2020universal,saito2021ovanet,saito2018open,you2019universal}. 
However, over-reliance on deep features can impede model adaptation performance, as they have a strong bias towards global information like texture rather than the essential object information like shape~\cite{geirhos2018imagenet,hermann2020origins}, which is considered by humans as the most critical cue for recognition~\cite{landau1988importance}.
Fortunately, recent studies have demonstrated that vision transformer (ViT)~\cite{kolesnikov2021image} exhibits a stronger shape bias than Convolutional Neural Network (CNN)~\cite{naseer2021intriguing,tuli2021convolutional}. 
As shown in Fig.~\ref{shape_texture} right, we confirmed that such strong object shape bias is mainly attributed to the self-attention mechanism, verified in a similar way as \cite{geirhos2018imagenet}. 
Figure~\ref{attn_vis1} demonstrates the attention vectors of samples accross domains. 
Although we can leverage the attention to focus on more object parts,  the
\textit{attention mismatch} problem may still exist due to domain shift, which refers to the attention vectors of same-class samples from different domains having some degree of the difference caused by potential variations in object size, orientation, and position across different domains. 
{Attention mismatch} can hinder the accurate classification of samples, especially when objects of different classes share similar sizes or positions. For example, in Figure~\ref{attn_vis1}, the kettle in the source domain and the flower in the target domain have more similar attention patterns.
Therefore, 
the key challenge in utilizing attention is to effectively explore and leverage the object information embedded in attention while mitigating the negative impact of attention mismatch.

In this paper, we propose a novel Universal Attention Matching (UniAM) framework to address the UniDA problem by leveraging both the feature and attention information in a complementary way. Specifically, UniAM introduces a Compressive Attention Matching (CAM) approach to solve the attention mismatch problem implicitly by sparsely representing target attentions using source attention prototypes. This allows CAM to identify the most relevant attention prototype for each target sample and distinguish irrelevant private labels.
Furthermore, a residual-based measurement is proposed in CAM to explicitly distinguish common and private samples across domains. 
By integrating attention information with features, we can mitigate the interference caused by domain shift and focus on label shift to some extent.
With the guidance of CAM, the UniAM framework achieves domain-wise and category-wise common feature alignment (CFA) and target class separation (TCS). 
By using an adversarial loss and a source contrastive loss, CFA identifies and aligns the common features across domains, ensuring their consistency and transferability.
On the other hand, TCS enhances the compactness of the target clusters, leading to better separation among all target classes. This is accomplished through a target contrastive loss, which encourages samples from the same target class to be closer together and farther apart from samples with other classes.



\textbf{Main Contributions}: (1) We propose the UniAM framework that comprehensively considers both attention and feature information, which allows for more accurate identification of common and private samples. (2) We validate the strong object bias of attention in ViT. To the best of our knowledge, we are the first to directly utilize attention in ViT for classification prediction.  (3) We implicitly explore object information by sparsely reconstructing attention, enabling better common feature alignment (CFA) and target class separation (TCS). (4) We conduct extensive experiments to show that UniAM can outperform current state-of-the-art approaches.

\section{Related Works}
\subsection{Universal Domain Adaptation}

UniDA~\cite{you2019universal} does not require prior knowledge of label set relationship. 
To address this problem, UAN ~\cite{you2019universal} proposes a criterion based on entropy and domain similarity to quantify sample transferability. CMU~\cite{fu2020learning} follows this paradigm to detect open classes by setting the mean of three uncertain scores including entropy, consistency and confidence as a new measurement. 
Afterward, \cite{kundu2020universal} proposes a real-time adaptive source-free UniDA method. 
In~\cite{saito2020universal} and ~\cite{li2021domain},  clustering is developed to solve this problem. 
~\cite{lifshitz2020sample}. OVANet~\cite{saito2021ovanet} employs a One-vs-All classifier for each class and decides known or unknown by using the output.
Recent works have shifted their focus towards finding mutually nearest neighbor samples of target samples~\cite{chen2022mutual, chen2022evidential, chen2022geometric, }  or constructing relationships between target samples and source prototypes~\cite{kundu2022subsidiary, chang2022unified}.


\begin{figure*}[tbp]
	\centering
	\subfloat
	{
        \includegraphics[width=6.8in]{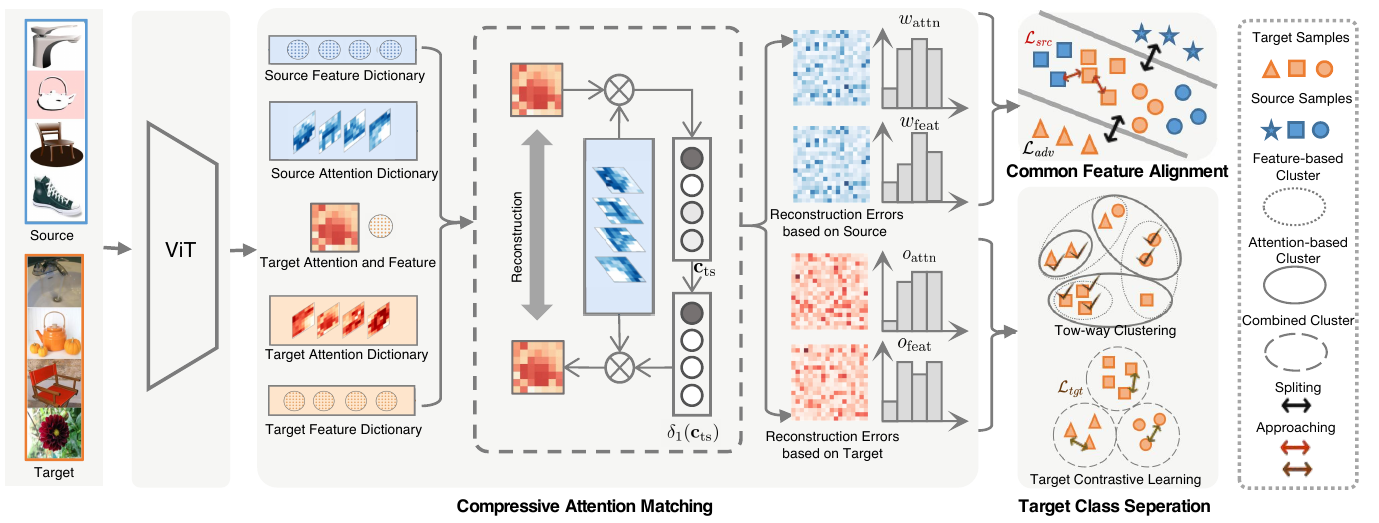}
	}
	\caption{\textbf{Illustration of the proposed UniAM framework.} The framework consists of  three integral components: Compressive Attention Matching (CAM), Common Feature Alignment (CFA) and Target Class Separation (TCS). 
  At its core, CAM reconstructs all target attentions and features based on the source dictionary (with feature reconstruction omitted in the figure for simplicity), and attention and feature commonness scores $w_\text{attn}$ and $w_\text{feat}$ are computed from residual vectors. 
  Then, domain- and category-wise CFA is achieved by minimizing $\mathcal{L}_{adv}$ and $\mathcal{L}_{src}$ guided by $w_\text{attn}$ and $w_\text{feat}$. 
  Similarly, $o_\text{attn}$ and $o_\text{feat}$ are obtained by reconstructing target attentions and features  based on the target dictionary in CAM. TCS performs two-way clustering from both the attention and feature views and minimizes $\mathcal{L}_{tgt}$ to achieve effective separation of target classes.}
	\label{framework}
\end{figure*}

\subsection{Vision Transformer}
Inspired by the success of Transformer~\cite{vaswani2017attention, khan2021pretrained} in the NLP field, many researchers have attempted to exploit it for solving computer vision tasks. One of the most pioneering works is Vision Transformer (ViT)\cite{kolesnikov2021image}, which decomposes input images into a sequence of fixed-size patches.
Different from CNNs that rely on image-specific inductive bias, ViT takes the advantage of large-scale pre-training data and global context modeling on the entire images.  Due to the outstanding performance of ViT, many approaches have been proposed based on it~\cite{touvron2021training,liu2021swin,wang2021pyramid,huang2023dbaformer,luo2022s2rl}, such as Touvron et al.~\cite{touvron2021training} propose DeiT, which introduces a distillation strategy specific to transformers to reduce computational costs. 
In general, ViT and its variants have achieved excellent results on many computer vision tasks, such as object detection ~\cite{carion2020end,zhu2020deformable,wang2021pyramid}, image segmentation ~\cite{zheng2021rethinking,wang2021end}, and video understanding ~\cite{girdhar2019video,neimark2021video}, etc. 

Recently, ViT has been adopted for the DA task in several works. TVT~\cite{yang2023tvt} proposes an transferable adaptation module to capture discriminative features and achieve domain alignment. SSRT~\cite{sun2022safe} formulates a comprehensive framework, which pairs a transformer backbone with a safe self-refinement strategy to navigate challenges associated with large domain gaps effectively. CDTrans~\cite{xu2021cdtrans} designs a triple-branch framework to apply self-attention and cross-attention for source-target domain feature alignment. Differently, we focus in this paper on investigating attention mechanism's superior discriminability across different classes on universal domain adaptation. 

\subsection{Sparse Representation Classification}
Sparse Representation Classification (SRC)~\cite{wright2008robust} and Collaborative Representation Classification (CRC)\cite{zhang2011sparse}, along with their numerous extensions~\cite{liu2018joint,wu2014collaborative,chi2013classification,chen2012dictionary,he2023prototype}, have been extensively investigated in the field of face recognition using single images and videos. These methods have demonstrated promising performance in the presence of occlusions and variations in illumination.
By modeling the test data in terms of a sparse linear combination of a dictionary, SRC can capture non-linear relationships between features. Our UniAM is inspired by them but uses a novel measurement instead of a sparsity concentration index.



\section{Problem Formulation and Preliminary}
\subsection{Problem Formulation}
Denoting $\mathbb{X}$, $\mathbb{Y}$, $\mathbb{Z}$ as the input space, label space and latent space, respectively. Elements of  $\mathbb{X}$, $\mathbb{Y}$, $\mathbb{Z}$  are noted as $\bm{x}$, $y$ and $\bm{z}$. Let $P_s$ and $P_t$ be the source distribution and target distribution, respectively.
We are given a labeled source domain $\mathbb{D}_{s}= \{\bm{x}_{i}, {y}_{i})\}_{i=1}^{m}$ and an unlabeled target domain $\mathbb{D}_{t}=  \{\bm{x}_{i}\}_{i=1}^{n}$ are respectively sampled from $P_{s}$ and $P_{t}$, where $m$ and $n$ denote the number of samples of source and target domains, respectively. Denote $\mathbb{L}_{s}$ and $\mathbb{L}_{t}$ as the label sets of the source and target domains, respectively. Let $\mathbb{L}=\mathbb{L}_{s} \cap \mathbb{L}_{t}$ be the common label set shared by both domains, while $\overline{\mathbb{L}}_{s}=\mathbb{L}_{s} \backslash \mathbb{L}$ and $\overline{\mathbb{L}}_{t}=\mathbb{L}_{t} \backslash \mathbb{L}$ be the label sets private to source and target domains, respectively. Denote $M = |\mathbb{L}_{s}|$ as the number of source labels.
Universal domain adaptation aims to predict labels of target data in $\mathbb{L}$ while rejecting the target data in $\overline{\mathbb{L}}^{t}$ based on $\mathbb{D}_{s}$ and $\mathbb{D}_{t}$. 

Our overall architecture consists of a ViT-based feature extractor, an adversarial domain classifier, and a label classifier. Suppose the function for learning embedding features is $G_f: \mathbb{X} \rightarrow \mathbb{Z} \in \mathbb{R}^{d_z}$ where $d_z$ is the length of each feature vector, the discrimination function of the label classifier is $G_c: \mathbb{Z} \rightarrow  \mathbb{Y} \in \mathbb{R}^{M}$, and the function of the domain classifier is $G_d: \mathbb{Z} \rightarrow \mathbb{R}^{1}$.


\subsection{Preliminary}

To start with, we provide an overview of the self-attention mechanism used in ViT. 
First, the input image $\bm{x}$ is divided into $N$ fixed-size patches, which are linearly embedded into a sequence of vectors. Next, a special token called the class token is prepended to the sequence of image patches for classification. The resulting sequence of length $N+1$  is then projected into three matrices: queries $\bm{Q}  \in \mathbb{R}^{(N+1) \times d_{k}}$, keys $\bm{K}  \in \mathbb{R}^{(N+1) \times d_{k}}$ and values $\bm{V}  \in \mathbb{R}^{(N+1) \times d_{v}}$  with $d_k$ and $d_v$ being the length of each query and value vector, respectively. Then, $\bm{Q}$ and $\bm{K}$ are passed to the self-attention layer to compute the patch-to-patch similarity matrix $\bm{A}^{(N+1) \times (N+1)}$, 
which is given by
\begin{equation}
      \bm{A}=\frac{\bm{Q} \bm{K}^\top}{\sqrt{d_k}},
      \label{attention}
\end{equation}
For ease of further processing, we flatten $\bm{A}$ into a vector $\bm{a} \in \mathbb{R}^{(N+1)^2 \times 1}$.
It is worth noting that multiple attention heads are utilized in the self-attention mechanism. Each head outputs a separate attention, and the final attention is obtained by concatenating the vectors from all heads. 
As a result, the dimensionality of $\bm{a} \in \mathbb{R}^{d_a \times 1}$, where $d_a = N_H \times (N+1)^2$ and $N_H$ is the number of attention heads. 
The utilization of multiple heads allows the model to jointly attend to information from different feature subspaces at different positions. 

Once the attention vector $\bm{a}$ is available, the corresponding $k$-th attention prototype $\bm{p}_k$ is calculated by averaging all attention vectors of samples in class $k$, which will be used in the subsequent matching process.

\section{Proposed Methodology}
\subsection{Compressive Attention Matching}\label{sec_cam}


Since the attention mismatch problem exists due to domain shift mentioned in Section~\ref{sec:intro}, 
how to effectively utilize the core object information and avoid interference from redundant information poses a challenge in applying attention to UniDA.
To address this challenge, compressive attention matching (CAM) is proposed to capture the most informative object structures by sparsely representing target attentions.
Define the attention dictionary in CAM as the collection of source attention prototypes for efficient matching, i.e.,  $\bm{P}_s=[\bm{p}^s_1, \bm{p}^s_2, \cdots,\bm{p}^s_{M}] \in \mathbb{R}^{d_a \times M}$. Definition~\ref{def_CAM} gives the definition of CAM.


\begin{definition} [\textbf{Compressive Attention Matching}]\label{def_CAM}
  Given an attention vector $\bm{a}_t  \in \mathbb{R}^{d_a \times 1}$ of the target sample $\bm{x}_t$ and a source attention dictionary $\bm{P}_s$, 
  Compressive Attention Matching aims to match $\bm{a}_t$  with one prototype in $\bm{P}_s$  to determine its commonness, which is achieved by assuming that $\bm{a}_t$ can be approximated by a linear combination of $\bm{P}_s$:
  \begin{equation}
   \bm{a}_t = {\bm{P}}_s \bm{c}_{ts},
   \label{cam}
  \end{equation}
  where the coefficient vector $\bm{c}_{ts} \in \mathbb{R}^{M \times 1}$ satisfies a \textbf{sparsity constraint} in order to achieve a compressive representation. Based on $\bm{c}_{ts}$, $\bm{x}_t$ is regarded as belonging to common classes from an attention perspective when the following inequality is satisfied:
  $$w_\text{attn}( \bm{x}_{t}) < \delta.$$
  $w_\text{attn}(\cdot)$ indicates a measurement to evaluate the commonness of $\bm{x}_t$ which is defined later and $\delta$ is a threshold.
\end{definition}
\noindent
\textbf{Why Compressive Attention Matching is desirable?} 
By enforcing sparsity on the coefficients in CAM, we can obtain a compressive representation of the attention vectors, which facilitates the extraction and utilization of low-dimensional structures embedded in high-dimensional attention vectors.
In the context of UniDA, this compressive representation enables us to identify the most relevant attention prototype for each target sample and distinguish irrelevant private labels, which is crucial for achieving effective common and private class detection. Therefore, CAM with sparse coefficients plays a vital role in solving UniDA.

To solve Eq.~\ref{cam} in CAM, the coefficient vector $\bm{c}_{ts}$ is estimated by:
  \begin{equation}
  \text{min}_{\bm{c}_{ts}}\|\bm{a}_t -\bm{P}_s{\bm{c}_{ts}} \|_2^2+\rho \|{\bm{c}_{ts}}\|_1,
  \label{optim_src}
  \end{equation} 
  where $\|\cdot\|_p$ denotes $\ell_p$-norm.
The $\ell_1$-minimization term in Eq. \ref{optim_src} yields a sparse solution, which enforces that $\bm{c}_{ts}$ has only a small number of non-zero coefficients.

 Then we can compute the class reconstruction error vector $\bm{r}_{ts} \in \mathbb{R}^{M}$ for each target sample using the sparse matrix $\bm{c}_{ts}$. The $k$-th entry of ${\bm{r}_{ts}}$ can be represented:
\begin{equation}
  \bm{r}_{ts}(k) =\|\bm{a}_t -\bm{P}_s\delta_k(\bm{c}_{ts}) \|_2, \quad k=1, \ldots, M,
\end{equation}
where $\delta_k(\bm{c}_{ts})$ is a one-hot vector with the $k$-th entry in $\bm{c}_{ts}$ being non-zero while setting all other entries to zero. 
If $\bm{x}_t$ corresponds to a common class $k$, then the reconstruction error corresponding to class $k$. $\bm{r}_{ts}(k)$ should be much lower than that corresponding to the other classes. Conversely, if $\bm{x}_t$ belongs to a private class, the difference between elements of the entire reconstruction error vector $\bm{r}_{ts}$ should be relatively small, without a significant difference between the errors corresponding to different classes.

As a result, the reconstruction error vector $\bm{r}_{ts}$ is a crucial component in CAM. 
It serves as the foundation for the design of the measurement $w_\text{attn}(\cdot)$ in Definition~\ref{def_CAM}  called Attention Commonness Degree (ACD), defined as belows:

\begin{definition} [\textbf{Attention Commonness Degree}]\label{acd} Given the residual vector $\bm{r}_{ts}$ of $\bm{x}_t$, the ACD is defined as the difference between the average of non-matched errors and matched errors:
  \begin{equation}
    w_\text{attn}(\bm{x}_t) =\text{non-match}({\bm{r}_{ts}}) - \text{match}({\bm{r}_{ts}}),
    \label{w_shape}
  \end{equation}
  where $\text{match}(\bm{r}_{ts}) = \bm{r}_{ts}(\hat{y})$,
  $\hat{y} =\arg \min _k \bm{r}_{ts}(k)$ and $\text{non-match}(\bm{r}_{ts})$  is the average of reconstruction errors excepting $\hat{y}$. 
\end{definition}
\begin{remark}
  ACD measures the degree of commonness for a target sample $\bm{x}_t$, which represents the probability of belonging to common classes. 
  A higher ACD value indicates a larger difference between non-matched and matched errors, suggesting the presence of an attention prototype similar to $\bm{x}_t$, and consequently, a higher degree of sample commonness. 
  Conversely, a smaller ACD value implies a similar reconstruction error between $\bm{x}_t$ and all source prototypes, indicating a lower degree of sample commonness and a higher degree of privateness.

\end{remark}

To complement the attention information, we retain features that reflect global information. The target feature $\bm{z}_t$ can be also represented by the linear span of source feature prototypes $\bm{Q}_s = [\bm{q}_1^s, \bm{q}_2^s,\cdots, \bm{q}_{M}^s]$, i.e.,$\quad \bm{z}_t = {\bm{Q}}_s{\bm{c}_{ts}}$. The corresponding residual vector $\bm{r}'_{ts}(k)$ is computed based on $\bm{c}_{ts}$. The Feature Commonness Degree (FCD) can be defined as
$
  w_\text{feat}(\bm{x}_t) =  \text{non-match}(\bm{r}'_{ts}) -  \text{match}(\bm{r}'_{ts}).
  \label{w_text}
$.

It is worth noting that by replacing $P_s$ in Definition \ref{cam} with the target dictionary $P_t$, we can obtain compressive representations of target attentions towards $P_t$. This leads to a score similar to that in Definition \ref{acd}, denoted as $o_\text{attn}$. The same goes for $o_\text{feat}$.  These scores can facilitate determining the probability that two target samples belong to the same class, more details will be provided in Section \ref{sec_tcp}.

In summary, both attention and feature characteristics are important factors that affect the perception of similarity between different categories. Attention captures the structural properties of objects, while feature captures the appearance properties of the global images. Therefore, we can achieve a more comprehensive and accurate private class detection model that takes into account both object information and global information.


\subsection{Common Feature Alignment}
To identify and align the common class features across domains, we propose a domain-wise and category-wise Common Feature Alignment (CFA) technique, which considers both attention and feature information.

\noindent
\textbf{Domain-wise Alignment.}
To achieve domain-wise alignment, we first propose a residual-based transferability score $d_\text{t}$ measuring the probability that the target sample belongs to the common classes, which can be summarized as:
\begin{equation}
  w_\text{t} = \lambda  w_\text{attn} + (1 - \lambda) w_\text{feat},
  \label{w_t}
\end{equation}
where $\lambda$ is a hyperparameter balancing their contribution.
Meanwhile, to measure the probability that the source sample $\bm{x}_s$ with label $j$ belongs to the common label set, we compute $w^s_j$ with the sum of all target samples' attention and feature reconstruction errors respectively, i.e.
\begin{equation}
  w_{s}^j =\lambda \sigma(\bm{r}_{ts})_j + (1 - \lambda)  \sigma(\bm{r}'_{ts})_j
\end{equation}
where $\bm{r}_{ts}^i$ indicates the reconstruction error of the $i$-th target sample. 
The operator $\sigma(\bm{r})=\frac{\bm{r} - min(\bm{r})}{max(\bm{r}) - min(\bm{r})}$ refers to the normalization sum of all target attention or feature residual vectors.
A larger value of $w^s_j$ indicates a higher probability that the source label $j$ belongs to the common label set, while lower values suggest that it is more likely to be a source private label. It is worth noting that source samples with the same label are assigned the same weight.

Based on the above two weights,  we can derive a domain-wise adversarial loss that aligns the common classes across domains as follows:
\begin{equation}
  \begin{aligned}
  \mathcal{L}_\text{adv} & =\mathbb{E}_{\bm{x}_s \in \mathbb{D}_S}\left[w_s \cdot \log \left(1-G_d\left(\bm{z}_s\right)\right)\right] \\
  & +\mathbb{E}_{\bm{x}_t \in \mathbb{D}_{T}}\left[w_t \cdot \log \left(G_d\left(\bm{z}_t\right)\right)\right],
  \end{aligned}
  \label{adv_loss}
\end{equation}
In addition, to avoid being interfered with the knowledge of source private samples, we employ an indicator as the weight for the weighted cross-entropy loss $\mathcal{L}_{\text{cls}}$ for the source domain, as shown below:
\begin{equation}
    \mathcal{L}_{\text{cls}} = -\mathbb{E}_{({\bm{x}_s},{y_s}) \in \mathbb{D}_S} \mathds{1}_{w^y_s \geqslant \alpha} l_{ce}({{y}}, G_c(G_f(\bm{x}_s))) ,
\end{equation}
where $l_{ce}$ is the standard cross-entropy loss. 



 
 


\noindent
\textbf{Category-wise Alignment.}
To enhance the source discriminability and align the common features from a category-wise perspective across domains, we propose a contrastive common feature alignment method. 
In order to quantify the likelihood that the source sample $\bm{x}_i^s$ and the sample $\bm{x}_j$ belong to the same category $y_i$, we design a category-wise target score $w_{i,j}$.
If $\bm{x}_j$ is a source sample, we use its ground truth label $y_j$ to determine if it's a positive or negative example, i.e., 
$
  w_{i,j} =\left\{\begin{array}{ll}
  1, \text { if } {y}_j=y_i  & \\
  0, \text { if } {y}_j \neq y_i  & \\
  \end{array}\right.
$.
If $\bm{x}_j$ is a target sample, we estimate the probability that it belongs to $y_i$ based on the residual vectors $\bm{r}_{ts}$ and $\bm{r}'_{ts}$. 
These two vectors can be seen as two vanilla prediction probability vectors and the corresponding pseudo-labels $\hat{y}_j$ and $\hat{y}'_j$ can be obtained by argmin operation. 
To give a more reliable estimation, the soft label of $\bm{x}_j$ is determined by these two pseudo-labels together.
Specifically, $w_{i,j}$ is set to $1$ when both $\hat{y}_j$ and $\hat{y}'_j$ are equal to $y_i$ and set to $0$ when both of them are not equal to $y_i$.
The soft label is computed as below when only one of the predictions is $y_i$:
\begin{equation}
  w_{i,j} =\left\{\begin{array}{ll}
    \lambda \cdot { w_\text{attn}}/{w_t}, 
     & \text{if } \hat{y}'_j  \neq  \hat{y}_j = y_i, \\
      (1-\lambda) \cdot w_\text{feat}/{w_t}, 
  & \text{if }  \hat{y}_j  \neq  \hat{y}'_j = y_i.  \\
  \end{array}\right.
  \label{w_ij}
\end{equation}


Thus the category-wise common feature alignment can be improved by minimizing the source contrastive loss $\mathcal{L}_{\text{src}}$:
\begin{equation}
\begin{aligned}
  & \mathcal{L}_{\text{src}} = -\mathbb{E}_{\bm{x}_i \in \mathbb{D}_S, \bm{x}_j \in \mathbb{D}_{S \cup T}} w_{i,j} l(\bm{z}_i,\bm{z}_j) \\ 
\end{aligned}
\end{equation}
with 
$$l(\bm{z}_i,\bm{z}_j) =\frac{\exp \left(\bm{z}_i \bm{z}_j / \tau\right)}{\sum_{k=1}^{m+n} \exp \left(\bm{z}_i \bm{z}_k / \tau\right)}.$$

\begin{table*}[htpb]
  \begin{center}
  \caption{\textbf{H-score (\%) on Office-31 and DomainNet}}
  \label{table:office_domainnet}
  \resizebox{\textwidth}{!}{
  \begin{tabular}{c|c|cccccccccccccc}
  \toprule[2pt]
  \multirow{2}{*}{Method} &
  \multicolumn{7}{c}{\textbf{Office-31}} &
  \multicolumn{6}{c}{\textbf{DomainNet}} 
  \\
  \cmidrule(r){3-9}\cmidrule(r){10-16} 
   & &
    \multicolumn{1}{c}{A2W} &
    \multicolumn{1}{c}{D2W} &
    \multicolumn{1}{c}{W2D} &
    \multicolumn{1}{c}{A2D} &
    \multicolumn{1}{c}{D2A} &
    \multicolumn{1}{c}{W2A} &
    \multicolumn{1}{c}{\textbf{Avg}} &
    \multicolumn{1}{c}{P2R} &
    \multicolumn{1}{c}{R2P} &
    \multicolumn{1}{c}{P2S} &
    \multicolumn{1}{c}{S2P} &
    \multicolumn{1}{c}{R2S} &
    \multicolumn{1}{c}{S2R} &
    \multicolumn{1}{c}{\textbf{Avg}} 
    \\  
  \midrule
  ResNet~\cite{he2016deep} &
  \multirow{11}{*}{\rotatebox{90}{ResNet50}}  & 47.92    & 54.94   & 55.60 & 49.78  & 48.48    & 48.96   & 50.94  & 30.06 & 28.34 & 26.95 & 26.95 & 26.89 & 29.74 & 28.15                \\
  DANN~\cite{ganin2016domain}  &      & 48.82    & 52.73    & 54.87     & 50.18      & 47.69        & 49.33   & 50.60  & 31.18 & 29.33 & 27.84 & 27.84 & 27.77 & 30.84 & 29.13                   \\
  OSBP~\cite{saito2018open}   &     & 50.23        & 55.53   & 57.20    & 51.14      & 49.75       & 50.16    & 52.34     & 33.60 & 33.03 & 30.55 & 30.53 & 30.61 & 33.65 & 32.00                \\
  UAN~\cite{you2019universal}  &     & 58.61         & 70.62      & 71.42    & 59.68        & 60.11     & 60.34  & 63.46   & 41.85 & 43.59 & 39.06 & 38.95 & 38.73 & 43.69 & 40.98                \\
  CMU~\cite{fu2020learning}   &     & 67.33       & 79.32         & 80.42       & 68.11       & 71.42           & 72.23    & 50.78 & 52.16 & 45.12 & 44.82 & 45.64 & 50.97 & 48.25 & 73.14                  \\
  DCC~\cite{li2021domain}  &       & 78.54      & 79.29       & 88.58          & 88.50      & 70.18        & 75.87 & 80.16   & 56.90 & 50.25 & 43.66 & 44.92 & 43.31 &  56.15 & 49.20                 \\ 
  OVANet~\cite{saito2021ovanet} &   & 79.45  & {95.43} & {94.35}   & 85.67  & 80.43   & 84.23 & 86.59  & 56.0 & 51.7 & 47.1 & 47.4 & 44.9 & 57.2 & 50.7         \\ 
    {UniOT~\cite{chang2022unified}} &    &   {89.16}  &      {98.93}  &   {96.87}     &    {86.35}   & 
    {89.85}      &      {88.08}  &
    {91.54}  & 59.30   &  47.79 & 51.79  & 46.81  &  48.32   & 58.25   & 52.04 
  \\ 
  \midrule
  $\text{OVANet}^{\star}$ & 
  \multirow{3}{*}{\rotatebox{90}{ViT}}    &    87.75     &     93.14     &       85.72   &    82.96     & \textbf{92.67}           & 91.25        &  88.92  &   {71.24} & 61.14 & 51.28 & 55.30 & 47.51  & 66.48   & 58.83             \\
  $\text{UniOT}^{\star}$ & 
   &   {96.35}    &     {99.13}       &    {99.43}      &   {88.40}      &      {89.67}     &   \textbf{93.81}   &
    {94.47} &  72.40  & 59.47  &  49.30 &  56.86 &  47.38   &  69.43 & 59.14 \\ \rowcolor{mygray}
  
   {Ours} &  &\textbf{95.46} & \textbf{99.62} & \textbf{99.81} & \textbf{95.28} & {92.35} & {93.23} & \textbf{95.95}  & \textbf{73.87} & \textbf{60.89}  & \textbf{52.31} &  \textbf{59.98} & \textbf{51.41}  & \textbf{70.68} & \textbf{61.52} 

  \\ 
  \bottomrule[2pt]
  \end{tabular}}
  \end{center}
  \end{table*}

\begin{table*}[htpb] 
  \begin{center}
  \caption{\textbf{H-score (\%) on Office-Home and VisDA2017}}
  \label{table:officehome_visda}
  \resizebox{\textwidth}{!}{
  \begin{tabular}{c|c|cccccccccccccc}
  \toprule[2pt]
  \multirow{3}{*}{Method} & &
  \multicolumn{13}{c}{\textbf{Office-Home}}&
  \multicolumn{1}{c}{\textbf{VisDA}}\\
  \cmidrule{3-15}\cmidrule{16-16}
  
  & & Ar2Cl & Ar2Pr & Ar2Rw & Cl2Ar & Cl2Pr & Cl2Rw & Pr2Ar & Pr2Cl & Pr2Rw & Rw2Ar & Rw2Cl & Rw2Pr & \textbf{Avg} & \textbf{S2R}\\ 
  \midrule
  ResNet~\cite{he2016deep} & \multirow{11}{*}{\rotatebox{90}{ResNet50}} & 44.65 & 48.04 & 50.13 & 46.64 & 46.91 & 48.96 & 47.47 & 43.17 & 50.23 & 48.45 & 44.76 & 48.43 & 47.32&{25.44} \\
  DANN~\cite{ganin2016domain} &   & 42.36 & 48.02 & 48.87 & 45.48 & 46.47 & 48.37 & 45.75 & 42.55 & 48.70 & 47.61 & 42.67 & 47.40 & 46.19&{25.65} \\
  OSBP~\cite{saito2018open}  &  & 39.59 & 45.09 & 46.17 & 45.70 & 45.24 & 46.75 & 45.26 & 40.54 & 45.75 & 45.08 & 41.64 & 46.90 & 44.48&{27.31} \\
  UAN~\cite{you2019universal} &    & 51.64 & 51.70 & 54.30 & 61.74 & 57.63 & 61.86 & 50.38 & 47.62 & 61.46 & 62.87 & 52.61 & 65.19 & 56.58&{30.47} \\
  CMU~\cite{fu2020learning} &    & 56.02 & 56.93 & 59.15 & 66.95 & 64.27 & 67.82 & 54.72 & 51.09 & 66.39 & 68.24 & 57.89 & 69.73 & 61.60&{34.64} \\
  DCC~\cite{li2021domain} &    & 57.97 & 54.05 & 58.01 & 74.64 & 70.62 & 77.52 & 64.34 & \textbf{73.60} & 74.94 & 80.96 & \textbf{75.12} & 80.38 & 70.18&{43.02} \\ 
  OVANet~\cite{saito2021ovanet}  &   & 62.81   & 75.54  & 78.59  & 70.72   & 68.78  & 75.03 & 71.27 & 58.64 & 80.52 &  76.09 & 64.13 & 78.91   &  71.75& {53.10}  \\ 
  {UniOT~\cite{chang2022unified}} &    &{67.27}  &   {80.54}  &      {86.03}   &      {73.51} &   
    {77.33}   & 
    {84.28}  &    
    {75.54}   &  
    {63.33}  &  
    {85.99} & 
    {77.77}      &     {65.37}     &      {81.92}    & {76.57}&{57.32}
  \\ 
  \midrule
  $\text{OVANet}^{\star}$ & \multirow{3}{*}{\rotatebox{90}{ViT}} &
    {58.09} & {86.06 } & {89.38} & {81.86} & {81.03} & {86.22} & {84.49} & {57.06} & {88.54} & {83.67} & {57.32} & {86.67} & {77.45}&{56.98} \\ 
    $\text{UniOT}^{\star}$ &  & {63.77} & \textbf{88.19} &      {90.23}   & {74.99} & {81.02} & 
    {84.55}  &{78.91}   & {61.29}  &  {87.60} & {82.38}&{63.70}&{88.30}&{78.40} &{63.25}
  \\ 
  \rowcolor{mygray} 
  Ours &  &
    \textbf{72.04} & {87.07} & \textbf{90.67} & \textbf{80.30} & \textbf{82.39} & {79.81} & \textbf{85.02} & {68.35} & \textbf{88.98} & \textbf{85.44} & \textbf{72.11} & {86.12} & \textbf{81.68} & \textbf{65.18} \\
    \bottomrule[2pt]
  \end{tabular}}
  \end{center}
  \end{table*}

\subsection{Target Class Seperation}\label{sec_tcp}
To better distinguish the common and private classes in the target domain, we propose a Target Class Separation (TCS) technique.
As mentioned in Section \ref{sec_cam}, we can construct a CAM problem based on the target attention dictionary $\bm{P}_t$.
As the target labels are unknown, $\bm{P}_t = [\bm{p}^t_1, \bm{p}^t_2,$ $\cdots,\bm{p}^t_{K}] \in \mathbb{R}^{d_a \times K}$ is initialized by performing traditional K-means algorithm on target attentions $\{\bm{a}^i_t\}_{i=1}^{n}$, where $K$ is pre-defined. In the subsequent attention clustering process, we calculate the residual vector $\bm{r}_{tt} \in \mathbb{R}^K$ and use it as a metric for measuring the distance between samples, which allows for dynamically update $\bm{P}_t$. Iteratively refining $\bm{P}_t$ makes it more reliable and discriminative. 
Meanwhile, the feature clustering based on the target feature dictionary  $\bm{Q}_t =  [\bm{q}^t_1, \bm{q}^t_2,$ $\cdots,\bm{q}^t_{K}] \in \mathbb{R}^{d_z \times K}$ is also performed. 
After the two-way clustering, each target sample $\bm{x}^t_i$ is assigned two cluster indexes $\hat{c}_i$ and $\hat{c}_i'$ from attention and feature view like Fig.~\ref{} depicted.  
The final soft pseudo label $o_{c, i}$ determining whether $\bm{x}_i$ belong to the $c$-th cluster is obtained based on these two cluster indexes, similar to $w_{i,j}$. 
Based on $o_{i,j}$, the target contrastive loss $\mathcal{L}_\text{tgt}$ is computed as follows:
\begin{equation}
  \begin{aligned}
    & \mathcal{L}_{\text{tgt}} = -\mathbb{E}_{\bm{x}_i \in \mathbb{D}_T, \bm{x}_j \in \mathbb{D}_{T}} o_{i,j} l(\bm{z}_i,\bm{z}_j), \\ 
  \end{aligned}
  \end{equation}
  with 
  $$l(\bm{z}_i,\bm{z}_j) =\frac{\exp \left(\bm{z}_i \bm{z}_j / \tau\right)}{\sum_{k=1}^{n} \exp \left(\bm{z}_i \bm{z}_k / \tau\right)}.$$
where $o_{i,j} = o_{c,i} \cdot o_{c,j}$ is the probability weight determining whether $\bm{x}_i$ and $\bm{x}_j$ belong to the same cluster $c$.
By minimizing $\mathcal{L}_{\text{tgt}}$, we can enhance the compactness of target clusters making a better separation among target classes.

\subsection{Overall Framework}



Overall, our framework is jointly optimized with four terms, i.e., cross-entropy loss $\mathcal{L}_{\text{cls}}$, adversarial loss $\mathcal{L}_{\text{adv}}$, source and target contrastive loss $\mathcal{L}_\text{src}$ and $\mathcal{L}_\text{tgt}$ as shown in Fig.~\ref{framework},
\begin{equation}
\begin{aligned}
    & \max_{G_d} \min_{G_f, G_c} \mathcal{L}_{\text{cls}} + \eta_1 \mathcal{L}_\text{src} + \eta_2 \mathcal{L}_\text{tgt}  - \mathcal{L}_{\text{adv}}, \\
\end{aligned}  
\end{equation}
where $\eta_1$ and $\eta_2$ are set as 0.5 to balance each loss component.
In the testing phase, given each input target sample ${\bm{x}_t}$, we compute $w_{t}$ in \eqref{w_t}. For those samples that satisfy $w_{t} < \beta$ are assigned with the predicted source class, where $\beta$ is a validated threshold. Otherwise, the samples are marked as unknown.

\begin{figure*}[tbp]
	\centering

	\subfloat[]{
	\includegraphics[width=3.2in]{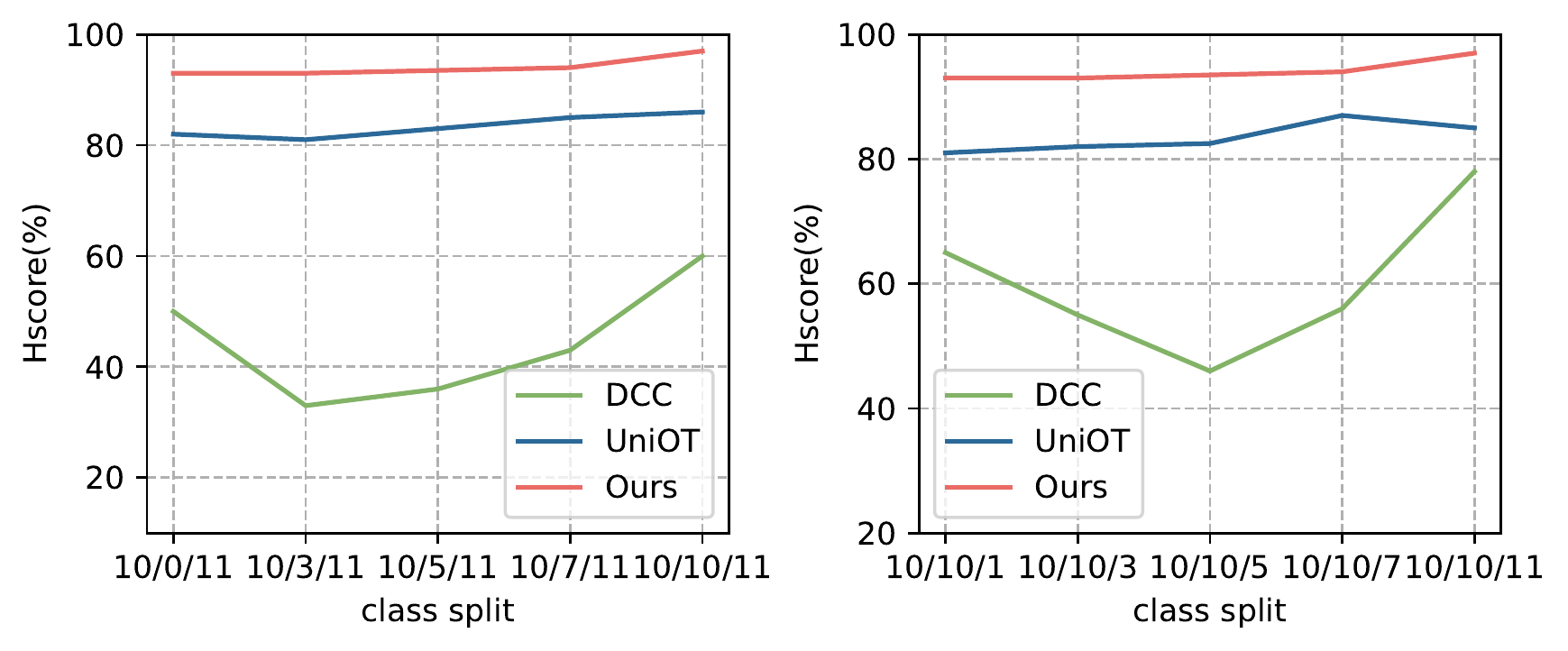}
	}
  \subfloat[]{
	\includegraphics[width=3.7in]{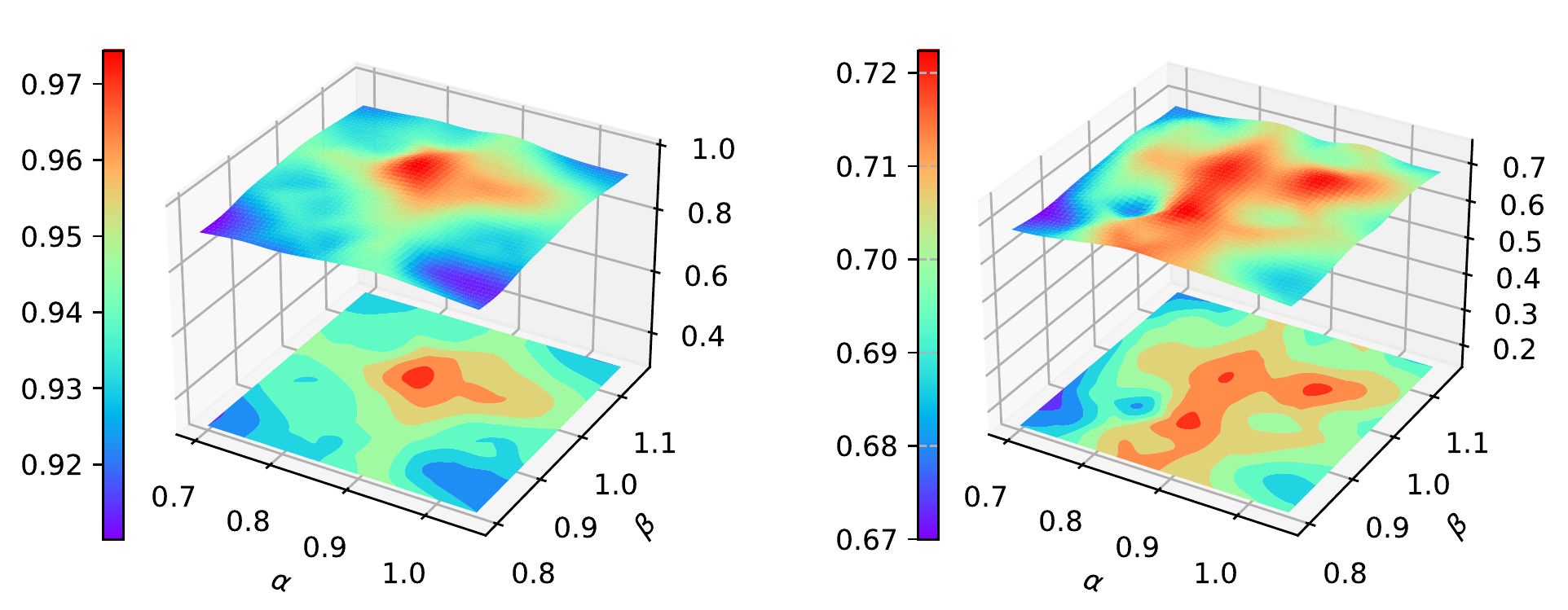}
	}
	\caption{(a) \textbf{Effectiveness on different label set relationships.} Our method consistently outperforms all comparative approaches across various settings of $ \overline{\mathbb{L}}^{t} $. (b) \textbf{Effectiveness of varying decision threshold $\alpha$ and $\beta$.} Even when both $ \alpha $ and $ \beta $ undergo significant fluctuations, the performance of our method doesn't decline sharply. 
	}
	\label{param_class_sen}
\end{figure*}


\begin{figure}[tbp]
	\centering
	\subfloat{
	\includegraphics[width=2.8in]{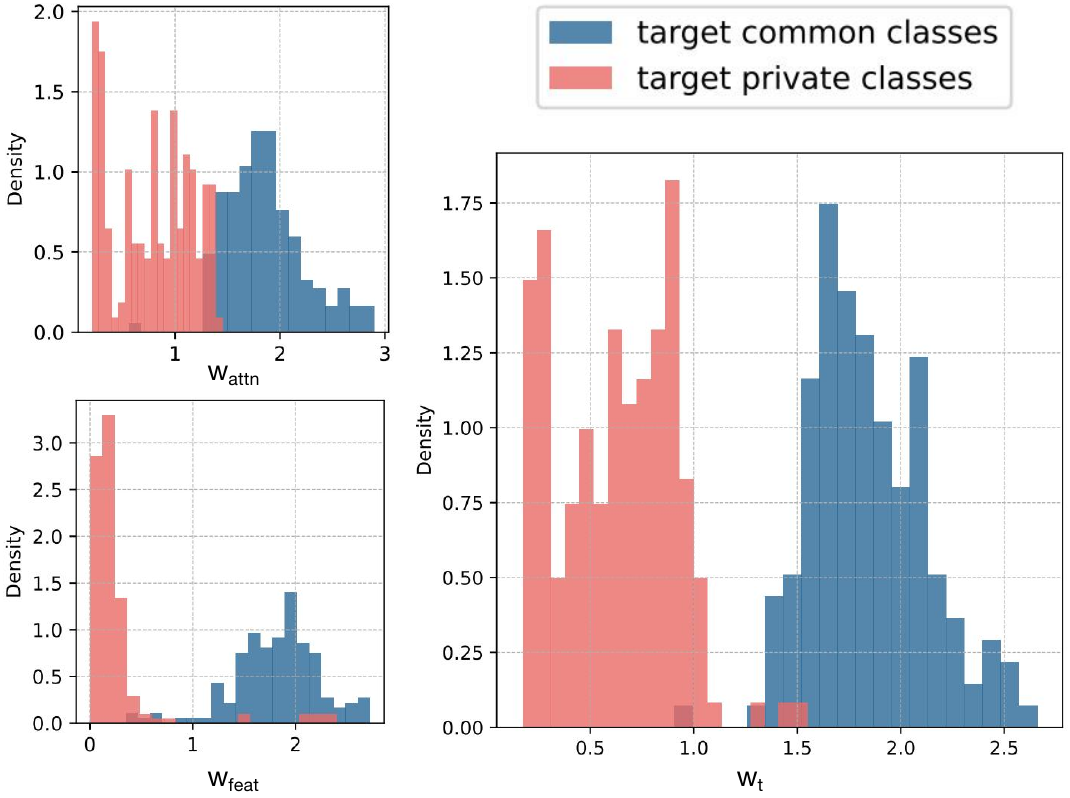}
	}
	\caption{\textbf{Qualitative analysis of feature and attention.} 
	}
	\label{criteria}
\end{figure}

\section{Experiment Results}

\subsection{Experimental Setup}
\noindent
\textbf{Datasets.}
We perform experiments on \textbf{Office-31} \cite{saenko2010adapting}, \textbf{Office-Home} \cite{venkateswara2017deep}, \textbf{VisDA2017} \cite{peng2018visda} and \textbf{DomainNet} \cite{peng2019moment} datasets. 
\textbf{Office-31} consists of three domains: Amazon (A), DSLR (D) and Webcam (W). Each  domain contains 31 categories. 
 \textbf{Office-Home} is a dataset made up of 65 different categories from four domains: Artistic (Ar), Clipart (Cl), Product (Pr) and Real-world images (Rw).
 \textbf{VisDA2017} is a dataset with a single source and target domain testing the ability to perform transfer learning from synthetic images to natural images. The dataset has 12 categories in each domain.  
We conduct experiments on three subsets from it, i.e., Painting (P), Real (R), and Sketch (S). For a fair comparison, we follow the same dataset split as \cite{you2019universal} for the first three dataset and \cite{fu2020learning} for the last dataset.


\noindent
\textbf{Evaluation Protocols.}
 We evaluate all methods using H-score~\cite{fu2020learning}. 
H-score is the harmonic mean of the accuracy of common classes and the accuracy of the ``unknown'' classes, which can make a trade-off between the accuracy of known and unknown classes. 


\noindent
\textbf{Implementation Details}
The method is implemented in Pytorch using a ViT-base model with $16 \times 16$ input patch size (or ViT-B/16) \cite{kolesnikov2021image}, pretrained on ImageNet~\cite{deng2009large}), as the backbone feature extractor. The transformer encoder of ViT-B/16 contains a total of 12 Transformer layers. The label classifier consists of a fully connected network with BatchNorm~\cite{ioffe2015batch}. The domain discriminator is a three-layer MLP with ReLU activations.
We train all models using a minibatch Stochastic Gradient Descent (SGD) optimizer with a momentum of $0.9$ and a weight decay of $5 \times 10^{-4}$. The learning rate decays by a factor of $(1+\alpha i / N )^{-\beta}$, where $i$ and $N$ respectively denote the current iteration and the global iteration. The batch size is set to 36. We initialize the initial learning rate to 0.01 for Office-31 and Office-Home, while set 0.001 for VisDA2017 and DomainNet. For the regularization hyperparameters, we set $\gamma = 100$ and $\lambda = 0.3$ for all dataset. For the decision threshold, we set $\alpha=0.85$ and $\beta=1.0$ for all dataset in UniDA and OSDA. In PDA, we set $\alpha=0.8$ for all dataset except $\alpha = 0.85$ in the Office-31 W2A task. For the pre-defined number of target prototypes, a larger size of the target domain indicates a larger $K$. Therefore, we empirically set $K = 50$ for Office-31, $K = 150$ for Office-Home, $K = 500$ for VisDA, $K = 1000$ for DomainNet.

\subsection{Comparison Baselines}

Follow the previous existing works~\cite{fu2020learning}, we compare our method with (1) ResNet~\cite{he2016deep}, (2) close-set domain adaptation: DANN~\cite{ganin2016domain}, (3) partial domain adaptation: PADA~\cite{cao2018partial}, ETN~\cite{cao2018partial1}, $\text{BA}^3\text{US}$~\cite{liang2020balanced} (4) open set domain adaptation: OSBP~\cite{saito2018open}, STA~\cite{liu2019separate}, ROS~\cite{bucci2020effectiveness}. (5)universal domain
adaptation: UAN~\cite{you2019universal}, CMU~\cite{fu2020learning}, DANCE~\cite{saito2020universal}, DCC~\cite{li2021domain}, OVANet~\cite{saito2021ovanet}, UniOT~\cite{chang2022unified}, GATE~\cite{chen2022geometric}.
We use some results from \cite{chen2022geometric}.
In all experiments, we assume that none of the UniDA methods have prior knowledge of category shift, while baselines tailored for each setting consider this prior.

\subsection{Comparison Results}
The experimental results for the Office-31, Office-Home, VisDA2017, and DomainNet datasets are presented in Table~\ref{table:office_domainnet}  and Table~\ref{table:officehome_visda}, which demonstrate that our proposed UniAM framework outperforms the state-of-the-art approaches in all benchmarks, as evaluated by the H-score metric.
Additionally, to ensure a fair comparison, we conducted experiments by replacing the backbone of OVANet and UniOT with ViT, marked as $\star $.  The proposed method consistently surpasses these ViT-based methods by a large margin. This indicates that our approach does not solely rely on using ViT as the backbone, but rather it fully exploits the advantage of the attention mechanism in ViT for UniDA tasks to achieve such superior performance.

\subsection{Analysis on Different Label Set Relationships}


\noindent
\textbf{Varying size of target private label set $\overline{\mathbb{L}}^{t}$.}
To explore the performance of our method under different class splitting settings with OVANet, UniOT and $\text{UniOT}^{\star}$, we fix $\mathbb{L}^{s}$, $\mathbb{L}$ and change $\overline{\mathbb{L}}^{t}$ on task A$\rightarrow$W in Office-31 dataset.
As shown in Fig.~\ref{param_class_sen} (a) left, our method consistently outperforms all comparison methods under different $\overline{\mathbb{L}}^{t}$, proving that our method is effective and robust for different $\overline{\mathbb{L}}^{t}$. 
As $\overline{\mathbb{L}}^{t}$ increases, meaning there are many open classes, our method outperforms other methods by a large margin, demonstrating that our method is superior in detecting open classes.

\noindent
\textbf{Varying size of common label set $\mathbb{L}$}
We fix $\mathbb{L}^{s}$ and $\mathbb{L}^{t}$ and varying $\mathbb{L}$ on task A$\rightarrow$W in Office-31 dataset. 
We let $\overline{\mathbb{L}}^{s}$, $\overline{\mathbb{L}}^{t}$ to keep 10 and 11 and vary $\mathbb{L}$ from 0 to 10 . 
In particular, all target data should be marked as ``unknown'' when the source and target domains do not overlap on label sets. 
As shown in Fig.~\ref{param_class_sen} (a) right, our method consistently outperforms previous methods on all sizes of $\mathbb{L}$, indicating that our method can detect open classes more effectively.

\begin{table}[!t]
  \begin{center}
  \caption{\textbf{Evaluation of the effectiveness of UniAM.} 
  }
  \label{losses}
  \resizebox{0.5\textwidth}{!}{
  \begin{tabular}{cccccccc}
  \toprule[2pt]
  \multirow{1}{*}{Method} &
   A2W & D2W & W2D & A2D & D2A & W2A & Avg  \\ 
  \midrule
  ${\text{w/ } \mathcal{L}_{adv}}$ &  92.12   &  93.37 &  99.49  & 93.54  &  88.32 & 92.77 & 93.27     \\
  ${\text{w/ } \mathcal{L}_{src}}$&  89.61   & 98.58  & 99.57   &  91.65  & 91.35  & 92.73 & 93.91    \\
  ${\text{w/ } \mathcal{L}_{tgt}}$ & 93.26    & 98.27  &  99.78  & 92.10  & 89.21 & 89.97 & 93.76     \\
  ${\text{w/o } \mathcal{L}_{adv}}$ &  94.69   &  98.10 &  99.78  & \textbf{96.12}  &  92.05 & 92.35 & 95.51    \\
  ${\text{w/o } \mathcal{L}_{src}}$ &  93.78   &  98.43 &  99.78  & 94.69  &  91.94 & 93.47 & 95.34      \\
  ${\text{w/o } \mathcal{L}_{tgt}}$ &  90.76   &  98.20 & 99.78   &  92.41 & 91.69  & 93.13 & 94.32  \\
  \midrule
  ${\text{w/o } w_\text{attn}}$ &  94.76   & 97.54  & 97.26   & 94.31  & 91.22  & 90.69 &  {94.30}    \\
  \midrule
  \rowcolor{mygray}
  Ours  &
    \textbf{95.46} & \textbf{99.62} & \textbf{99.81} & {95.28} & \textbf{92.35} & \textbf{93.23} & \textbf{95.95}  \\
  \bottomrule[2pt]
  \end{tabular}
  }
  \end{center}
  \end{table}



\subsection{Analysis on Our Method}

\noindent
\textbf{{Effectiveness of different losses.}} 
As there are three losses excluding classification loss in our method, we conduct another experiment to verify the effectiveness of each loss and any combination of them on Office-31 dataset. 
As shown in the first six rows of Table.~\ref{losses}, 
the results indicate that the use of any single loss function or a combination of any two loss functions can lead to a decrease in performance to some extent, with a performance drop of 2.05\%-2.68\% observed when using a single loss. These findings demonstrate the effectiveness of our proposed method. 

\begin{figure*}[tbp]
	\centering
	\subfloat{
	\includegraphics[width=7in]{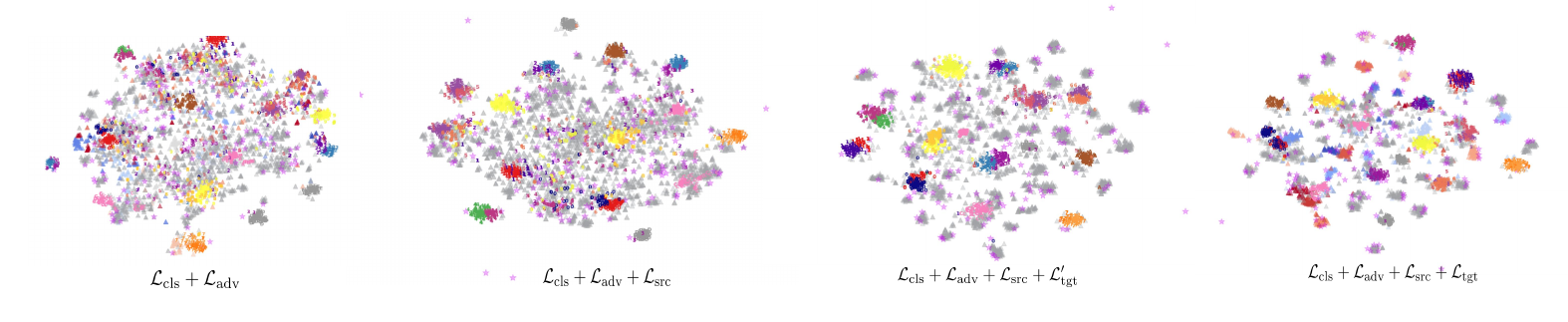}
	}
	\caption{\textbf{Feature visualization of target domain with different losses.} $\mathcal{L}'_\text{tgt}$ and $\mathcal{L}_\text{tgt}$ refer to using only $w_\text{feat}$ to calculate the loss weight $w_{i,j}$ and using both $w_\text{attn}$ and $w_\text{feat}$ together to calculate $w_{i,j}$. (a-c) $ \mathcal{L}\text{src} $ increases the distance between common and private categories, treating all target-private samples as one class, while $ \mathcal{L}\text{tgt} $ enhances discriminability of target private classes. (d) UniAM leverages attention to guide the refinement of target representations.}
	\label{tsne}
\end{figure*}

\noindent
\textbf{{Effectiveness of attention in ViT.}} 
To demonstrate the indispensable roles of attention in ViT in the proposed transferability criteria, we introduce a variant denoted as $\text{w/o } w_\text{attn}$, which performs sparse reconstruction only on the features.
Compared with our method in the last two rows of Table~\ref{losses}, the average performance drop of $\text{w/o } w_\text{attn}$  1.65\%. 
It indicates that the attention mechanism does play an effective role in attention enhancement on the basis of features during the process of common category detection. 

\noindent
\textbf{ Qualitative Analysis.} 
As shown in Fig.~\ref{criteria},
the three probability density histograms visualize partially $w_\text{attn}$, $w_{text}$, and their weighted sum $w_t$ in A$\rightarrow$W task on Office. From Fig.~\ref{criteria}, it can be observed that using  $w_\text{attn}$ or $w_{text}$ alone can partially distinguish common samples (colored in blue) and private samples (colored in red), but each has its limitations. $w_\text{attn}$ is prone to confusion at the boundary, while $w_{text}$ has some outliers, such as private samples with extremely high values and common samples with extremely low values. By combining them together, these two limitations can be effectively alleviated. The weighted sum $w_t$ can result in clearer boundaries between private and common samples, and the outliers are reduced.

\noindent
\textbf{Feature visualization.} We use t-SNE to visualize the learned target features for Pr$\rightarrow$Rw of Office-Home. As shown in Fig.~\ref{tsne}, the gray dots represent private samples, while the non-gray dots represent common samples, and their colors indicate their ground-truth classes.
Fig.~\ref{tsne} (a)-(c) shows that $\mathcal{L}_\text{src}$ increased the distance between common and private categories while all target-private samples are treated as a single class, and $\mathcal{L}_\text{tgt}$ improved the discriminability of the target private classes. 
Especially, Fig.~\ref{tsne} (d) validates that UniAM learns a better target representation introducing attention as a guide for attention enhancement can further improve the discriminability in the target domain by bringing same-class samples closer and pushing different-class samples farther away.

\noindent
\textbf{Sensitivity to decision threshold.}
We investigate the sensitivity of thresholds $\alpha$ and $\beta$, which are used to determine whether source and target samples belong to common classes respectively. The analysis was done in A$\rightarrow$D on Office-31 and Ar$\rightarrow$Cl on Office-Home.
As depicted in Fig.~\ref{param_class_sen} (b), the H-score demonstrates minimal variance. Specifically, $ \alpha $ varies within a reasonable and practical range of [0.7, 1.0], while $ \beta $ varies in a range of [0.8, 1.1]. These findings collectively reinforce the idea that our method remains robust to variations in the $ \alpha $ and $ \beta $ parameters. This robustness is a strong indicator of the method's stability and resilience under varying parameter settings.




\section{Conclusions}
In this work, we introduced UniAM, an innovative Compressive Attention Matching framework. 
What distinguishes UniAM is its unique capability to exploit the self-attention mechanism inherent in ViT, allowing it to adeptly capture the most pertinent information necessary for Universal Domain Adaptation. This is further complemented by its innovative compressive reconstruction module and residual-based transferability criterion, which together enable effective domain alignment. It's worth noting that UniAM stands as a pioneering method that directly harnesses the attention capabilities of vision transformers, specifically for classification tasks. 
Through extensive experiments on four benchmark datasets, we've found that our approach consistently eclipses the state-of-the-art UniDA method in both common set accuracy and "unknown" class accuracy.
We hope these findings will provide a new perspective for domain adaptation and other fields such as out of distribution detection in this future.

\section*{Acknowledgements}
This work was supported by the National Key Research and Development Project of China (2021ZD0110505), National Natural Science Foundation of China (U19B2042, 62006207, U20A20387), the Zhejiang Provincial Key Research and Development Project (2023C01043), University Synergy Innovation Program of Anhui Province (GXXT-2021-004), Academy Of Social Governance Zhejiang University, Fundamental Research Funds for the Central Universities (226-2022-00064, 226-2022-00142).

{\small
\bibliographystyle{ieee_fullname}
\bibliography{egbib}
}

\ificcvfinal\pagestyle{empty}\fi

\end{document}